\documentclass{INTERSPEECH2023}

\usepackage{xcolor}
\hypersetup{
    colorlinks,
    linkcolor={black},
    citecolor={black},
    urlcolor={black}
}
\usepackage{hyperref}
\usepackage{color}
\usepackage{booktabs}
\usepackage{amssymb,xcolor,stackengine,graphicx}
\usepackage{multirow}
\usepackage{textpos}
\definecolor{mygray}{gray}{0.6}
\interspeechcameraready 
\newcommand{\daniel}[1]{}

\newcommand{\wei}[1]{} 

\title{Improving Textless Spoken Language Understanding with Discrete Units as Intermediate Target \wei{an?}}
\name{Guan-Wei Wu$^{1\dagger}$, Guan-Ting Lin$^{1\dagger}$, Shang-Wen Li$^2$, Hung-yi Lee$^1$ \thanks{$^{\dagger}$Equal contribution.}}
\address{
  $^1$National Taiwan University, Taiwan \\
  $^2$Meta AI, USA}
\email{\{b08901019, f10942104, hungyilee\}@ntu.edu.tw \wei{other two authors' emails}}

\begin{document}

\maketitle

 
\begin{abstract}
Spoken Language Understanding (SLU) is a task that aims to extract semantic information from spoken utterances. Previous research has made progress in end-to-end SLU by using paired speech-text data, such as pre-trained Automatic Speech Recognition (ASR) models or paired text as intermediate targets. However, acquiring paired transcripts is expensive and impractical for unwritten languages. On the other hand, Textless SLU extracts semantic information from speech without utilizing paired transcripts. However, the absence of intermediate targets and training guidance for textless SLU often leads to suboptimal performance. In this work, inspired by the content-disentangled discrete units from self-supervised speech models, we proposed to use discrete units as intermediate guidance to improve textless SLU performance. Our method surpasses the baseline method on five SLU benchmark corpora. Additionally, we find that unit guidance facilitates few-shot learning and enhances noise robustness.







\end{abstract}



\section{Introduction}
Spoken Language Understanding (SLU) is a task that aims to capture semantic information from spoken utterances. Recent advancements in SLU have led to successful applications in various industries, such as voice assistants and voice-controlled smart devices. The traditional approach to SLU involves a cascaded system comprising an Automatic Speech Recognition (ASR) model followed by a text-based Natural Language Understanding (NLU) model. This approach has shown promising results, but training the ASR model requires a large amount of labeled data, which can be challenging and costly to obtain. Additionally, there are many languages that are primarily spoken and lack written sources, making it impossible to build an ASR system for these languages. Another research direction is \textit{end-to-end SLU}~\cite{serdyuk2018towards}, which intends to predict semantic labels directly from speech features. However, previous studies rely on using a pre-trained ASR model as model initialization or jointly training ASR/NLU and SLU simultaneously~\cite{saxon21_interspeech,multi-task, thomas2021rnn, arora2022espnet, wang2021speech2slot, thomas2022towards} with paired transcripts guidance.

To alleviate the reliance on paired transcripts, we step forward to \textit{Textless SLU}~\cite{kuo2020end}, which extracts the semantic information without the need for paired transcripts\footnote{Noting that if the task requires to predict entity names, such as slot filing, the entity names are still utilized for training, while all other parts of transcripts are discarded.}. Our approach draws inspiration from recent advancements in Self-Supervised Learning (SSL) of speech~\cite{SSL-review, superb, lin2023utility} and Textless NLP~\cite{textless_nlp_a, textless_nlp_b, textless_nlp_c, lin22c_interspeech, kharitonov-etal-2022-textless}. It has been observed that the self-supervised discovered units are highly related to content information~\cite{textless_nlp_b}. Based on this observation, we hypothesize that discrete units can effectively serve as unsupervised targets for capturing content information in SLU.

\begin{figure}{}
    \centering
\includegraphics[width=1.0\linewidth]{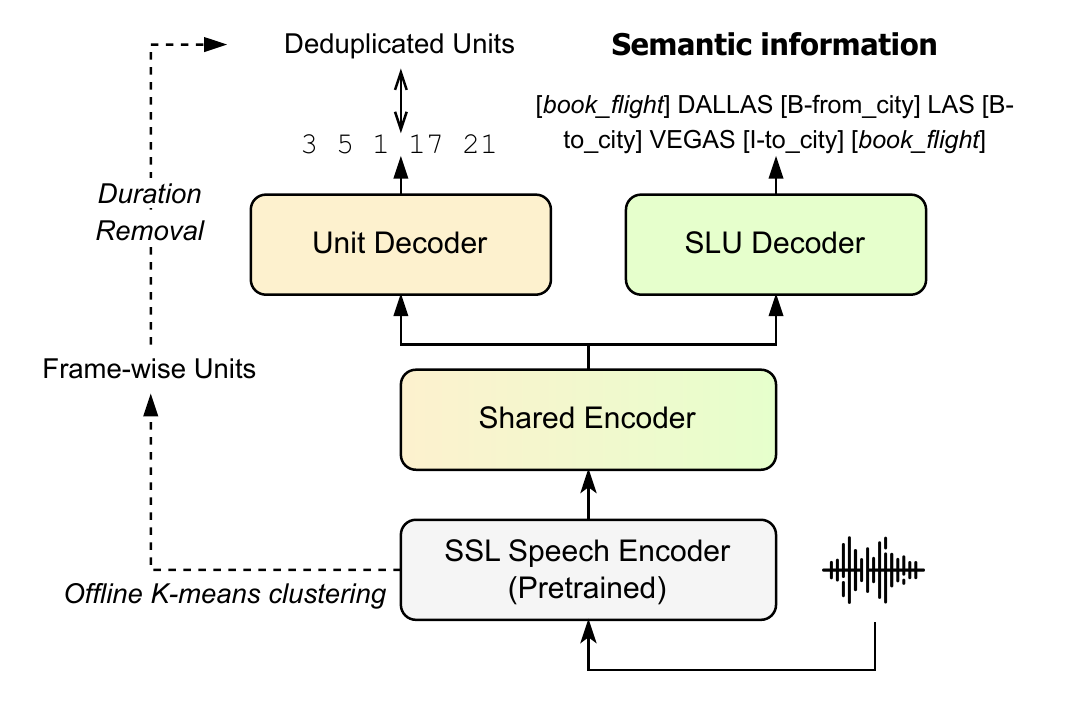}
    \vspace{-0.5cm}
    \caption{The proposed textless SLU framework with discrete units as the intermediate target. The transformer encoder is shared between the unit decoder and SLU decoder. Noting that we train the model to predict only semantic information without using full transcripts of the utterances.}
    \vspace{-0.5cm}
    \label{fig:framework}
\end{figure}

In this work, we propose to leverage self-supervised discovered speech units as the intermediate target for end-to-end textless SLU. Our approach involves incorporating a transformer-based encoder-decoder model for the main SLU task, along with a separate decoder that shares the same encoder to predict discrete units. The underlying rationale is that discrete units primarily encode content-related information. By using them as guidance, we encourage the intermediate representation to prioritize content information, which is crucial for SLU tasks.

To evaluate our approach, we conduct experiments on five publicly available SLU benchmarks: ATIS~\cite{atis}, SLUE-SNER~\cite{slue}, SLURP~\cite{slurp}, SNIPS~\cite{audio_snips}, and STOP~\cite{stop}. The experimental results demonstrate that our proposed method generally enhances textless SLU performance compared to the baseline and achieves competitive performance comparable to methods utilizing paired transcripts. For instance, on the ATIS dataset, we observe a 5.8\% improvement in F1-score and a 3.6\% improvement in intent accuracy compared to the approach by Morais \textit{et al.}~\cite{multi-task} under a similar architecture when not using text-speech paired data.

Moreover, our findings suggest that the proposed method exhibits robustness in few-shot and noisy scenarios. The model trained with unit guidance consistently outperforms the baseline approach in situations with limited data or domain mismatch. This indicates that our work can enhance the performance and generalizability of the model when faced with data scarcity or variations in domain. As discrete units predominantly contain content-related information, the shared encoder can prioritize content information by learning to predict these units. This focus on content information makes our method more robust to irrelevant noise. Furthermore, since SLU tasks aim to capture semantics from spoken content, unit prediction acts as a content-related regularization technique, thereby improving few-shot training.

The primary contributions of this paper are as follows: 1) Proposing self-supervised units as guidance for textless SLU. 2) Demonstrating consistent improvement on different SLU benchmark corpora. 3) Showing the proposed method's capacity for few-shot and noise robustness.




\section{Method}
\subsection{Problem Formulation}
Spoken Language Understanding (SLU) tasks intend to capture semantic information in an utterance. We consider three types of SLU tasks: 1) \textbf{Spoken Name Entity Recognition (SNER)}: Extract entity names and their corresponding labels. 2) \textbf{Joint Intent Classification and Slot Filling (ICSF)}: Similar to SNER, but also needs to recognize the utterance-wise intent label. 3) \textbf{Spoken Semantic Parsing (SSP)}: Creating a semantic parsing tree based on the utterance.

\wei{Whether the ``x'', ``N", ``C", ... are needed?}

\begin{figure}{}
    \centering
    \includegraphics[width=1.0\linewidth]{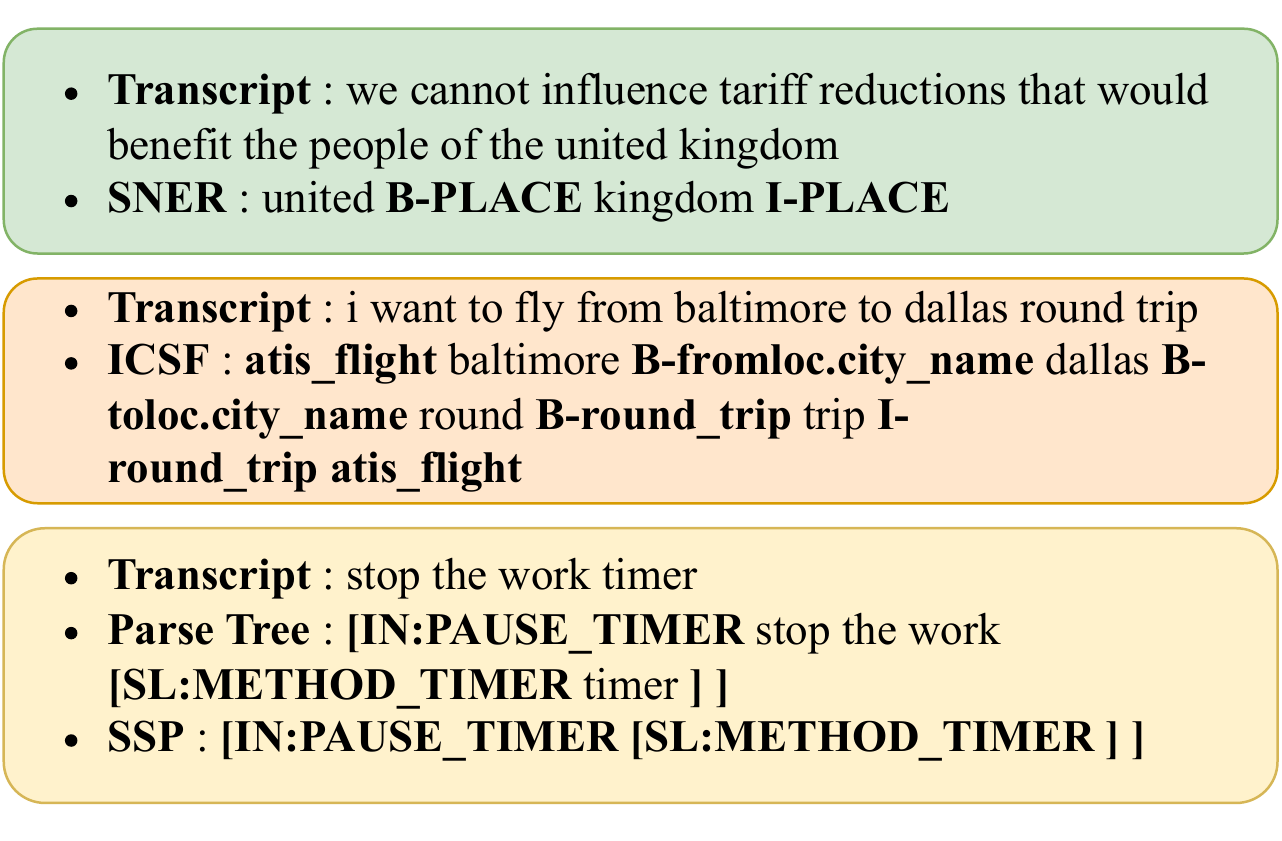}
    \vspace{-0.7cm}
    \caption{Transcriptions and sequence configurations for three SLU tasks, including SNER, ICSF, and SSP.}
    \vspace{-0.7cm}
    \label{fig:my_seq}
\end{figure}

We formulate these different textless SLU tasks as sequence generation tasks. For the SNER task, the targeted ground truth output sequence is constructed in the BI format, which means adding ``B-" as a prefix to the entity class name for the first word, and ``I-" for the rest of words in the entity. For the ICSF task, sequences are designed in a similar BI format as SNER, but with added intent class labels at the beginning and end of the sequence. Finally, for the SSP task, the sparse tree is transformed into a sequence format where different semantic labels are separated by brackets, along with the corresponding transcripts. This formulation helps maintain information on tree structures. However, due to the textless premise, we slightly modify the sequence from carrying the transcripts, which means that we only focus on predicting the semantic labels and the tree structure. Examples of each formulation are shown in Figure \ref{fig:my_seq}.

\wei{Whether the order of figure1 and figure2 is OK?\\figure1: ``K-means", the capitalization, boldface, and font size}

\begin{table}[t]
\caption{Evaluation metrics for different datasets.}
\vspace{-0.2cm}
\label{tab:dataset_metric}
\centering
\begin{tabular}{l|l}
\toprule
Dataset & Evaluation Metric \\ \midrule
ATIS & F1 / ST-F1 / INT-Acc \\ 
SLUE-SNER& F1 / ST-F1 / SV-CER \\
SLURP & SLU-F1 / INT-Acc \\ 
SNIPS & ST-F1 / SV-CER / INT-Acc \\
STOP & EM-Tree \\ \bottomrule
\end{tabular}
\vspace{-0.5cm}
\end{table}

\begin{table*}[t]
\caption{The performance of baseline, unit-guiding, and text-guiding approaches measured on test sets of ATIS, SLURP, SNIPS, STOP, and the development set of SLUE-SNER. The ``Previous" results are sourced from~\cite{multi-task} for ATIS,~\cite{slue} for SLUE-SNER,~\cite{arora2022espnet} for SLURP,~\cite{superb} for SNIPS, and~\cite{stop} for STOP. The notation ``$^*$" with \textcolor{mygray}{gray color} indicates that the training uses ASR transcripts, which are not directly comparable to our setup. ``N/A" means that we cannot find performance reported by prior works.}
\vspace{-0.3cm}
\label{tab:main_result}
\centering
\resizebox{0.98\textwidth}{!}{\begin{tabular}{l|ccc|ccc|cc|ccc|c}
\toprule
Dataset & \multicolumn{3}{c|}{\textbf{ATIS}}                            & \multicolumn{3}{c|}{\textbf{SLUE-SNER}}    & \multicolumn{2}{c|}{\textbf{SLURP}}    & \multicolumn{3}{c|}{\textbf{SNIPS}}    & \textbf{STOP} \\ \midrule
Metric & \multicolumn{1}{c}{F1$\uparrow$} & \multicolumn{1}{c}{ST-F1$\uparrow$} & INT-Acc$\uparrow$ & \multicolumn{1}{c}{F1$\uparrow$} & \multicolumn{1}{c}{ST-F1$\uparrow$} & SV-CER$\downarrow$ & \multicolumn{1}{c}{SLU-F1$\uparrow$} & INT-Acc$\uparrow$ & \multicolumn{1}{c}{ST-F1$\uparrow$} & \multicolumn{1}{c}{SV-CER$\downarrow$} & INT-Acc$\uparrow$ & EM-Tree$\uparrow$ \\ \midrule

Previous & \multicolumn{1}{c}{76.6} & \multicolumn{1}{c}{N/A} & 93.2 & \multicolumn{1}{c}{\textcolor{mygray}{70.3$^*$}} & \multicolumn{1}{c}{N/A} & N/A & \multicolumn{1}{c}{\textcolor{mygray}{71.9$^*$}} & 77.0 & \multicolumn{1}{c}{\textcolor{mygray}{89.8$^*$}} & \multicolumn{1}{c}{\textcolor{mygray}{21.8$^*$}} & N/A & \textcolor{mygray}{82.9$^*$}\\ 

Baseline & \multicolumn{1}{c}{79.1} & \multicolumn{1}{c}{84.3} & 96.5 & \multicolumn{1}{c}{64.8} & \multicolumn{1}{c}{74.1} & 35.2 & \multicolumn{1}{c}{63.2} & 78.7 & \multicolumn{1}{c}{77.6} & \multicolumn{1}{c}{42.9} & 96.7 & 80.0 \\ 
Unit & \multicolumn{1}{c}{\textbf{82.4}} & \multicolumn{1}{c}{\textbf{86.0}} & \textbf{96.8} & \multicolumn{1}{c}{\textbf{68.6}} & \multicolumn{1}{c}{\textbf{78.1}} & \textbf{29.4} & \multicolumn{1}{c}{\textbf{67.9}} & \textbf{80.9} & \multicolumn{1}{c}{\textbf{82.7}} & \multicolumn{1}{c}{\textbf{31.9}} & \textbf{97.0} & \textbf{84.4} \\ \midrule
Text & \multicolumn{1}{c}{84.5} & \multicolumn{1}{c}{87.7} & 97.4 & \multicolumn{1}{c}{69.2} & \multicolumn{1}{c}{78.2} & 29.0 & \multicolumn{1}{c}{69.9} & 82.5 & \multicolumn{1}{c}{83.2} & \multicolumn{1}{c}{30.7} & 97.0 & 84.5 \\ 

\bottomrule
\end{tabular}}
\end{table*}




\subsection{Self-Supervised Speech Units}

Recent studies in Textless NLP~\cite{textless_nlp_a, textless_nlp_b, textless_nlp_c, lin22c_interspeech, kharitonov-etal-2022-textless} have utilized self-supervised clustered discrete units to model speech and have shown that discrete units mostly preserve spoken content.
In this work, we propose to leverage these self-supervised units as intermediate guidance for content-related tasks like SLU. We follow a similar pipeline as Lin \textit{et al.}~\cite{lin22c_interspeech} to generate the discrete units. Specifically, we employ HuBERT-Large~\cite{hubert}, which is pre-trained on LibriLight~\cite{librilight} 60k-hour dataset, to produce 1024-dimension features. Each semantic frame representing 20 ms of speech. The resulting features are then passed through a K-means clustering model $(\rm{K}=500)$, which is pre-trained on LibriSpeech~\cite{librispeech} 100-hour dataset, to transform them into discrete tokens. Tokens are deduplicated to remove redundant information, resulting in shortened and simplified unit sequences (e.g., \texttt{17 17 5 8 8} to \texttt{17 5 8}) as the intermediate target for SLU.

\wei{italics? spoken content, Lin}

\begin{table*}[t]
\caption{The performance of few-shot training with 10\% of the training set on SLURP and SNIPS datasets. Values denoted in the $\delta$ column indicate the performance drop from 100\% training data to 10\% training data.}
\vspace{-0.3cm}
\label{tab:few_shot_result}
\centering
\resizebox{0.75\textwidth}{!}{\begin{tabular}{l|cccccc|cccccc}
\toprule
Dataset & \multicolumn{6}{c|}{\textbf{SLURP}}                                                                                                    & \multicolumn{6}{c}{\textbf{SNIPS}}                                                                                                    \\ \midrule
Metric & \multicolumn{3}{c|}{SLU-F1$\uparrow$}                                                 & \multicolumn{3}{c|}{INT-Acc$\uparrow$}                            & \multicolumn{3}{c|}{ST-F1$\uparrow$}                                                 & \multicolumn{3}{c}{SV-CER$\downarrow$}                            \\ \midrule
Portion & \multicolumn{1}{c}{100\%} & \multicolumn{1}{c}{10\%} & \multicolumn{1}{c|}{$\delta$} & \multicolumn{1}{c}{100\%} & \multicolumn{1}{c}{10\%} & $\delta$ & \multicolumn{1}{c}{100\%} & \multicolumn{1}{c}{10\%} & \multicolumn{1}{c|}{$\delta$} & \multicolumn{1}{c}{100\%} & \multicolumn{1}{c}{10\%} & $\delta$ \\ \midrule

Baseline & \multicolumn{1}{c}{63.2} & \multicolumn{1}{c}{45.2} & \multicolumn{1}{c|}{18.0} & \multicolumn{1}{c}{78.7} & \multicolumn{1}{c}{54.7} & 24.0 & \multicolumn{1}{c}{77.6} & \multicolumn{1}{c}{64.2} & \multicolumn{1}{c|}{13.4} & \multicolumn{1}{c}{42.9} & \multicolumn{1}{c}{62.2} & 19.3 \\
Unit & \multicolumn{1}{c}{\textbf{67.9}} & \multicolumn{1}{c}{\textbf{53.7}} & \multicolumn{1}{c|}{\textbf{14.2}} & \multicolumn{1}{c}{\textbf{80.9}} & \multicolumn{1}{c}{\textbf{69.5}} & \textbf{11.4} & \multicolumn{1}{c}{\textbf{82.7}} & \multicolumn{1}{c}{\textbf{78.0}} & \multicolumn{1}{c|}{\textbf{ 4.7}} & \multicolumn{1}{c}{\textbf{31.9}} & \multicolumn{1}{c}{\textbf{40.3}} & \textbf{ 8.4} \\
\bottomrule
\end{tabular}}
\vspace{-0.2cm}
\end{table*}

\begin{table*}[t]
\caption{Results of the performance drop on the test set of SLURP dataset under different noisy configurations. G is Gaussian noise with the followed amplitude value. M is MUSAN background noise, with the dB value representing signal-to-noise ratio. Reverb represents the reverberation effect.}
\vspace{-0.2cm}
\label{tab:noise_result_slurp}
\centering
\resizebox{0.98\textwidth}{!}{\begin{tabular}{l|cccccc|cccccc}
\toprule
Metric & \multicolumn{6}{c|}{SLU-F1$\uparrow$}                                                                                                                         & \multicolumn{6}{c}{INT-Acc$\uparrow$}                                                                                                    \\ \midrule
Noise & \multicolumn{1}{c}{w/o} & \multicolumn{1}{c}{G-0.01} & \multicolumn{1}{c}{G-0.02} & \multicolumn{1}{c}{M-20dB} & \multicolumn{1}{c}{M-10dB} & \multicolumn{1}{c|}{Reverb} & \multicolumn{1}{c}{w/o} & \multicolumn{1}{c}{G-0.01} & \multicolumn{1}{c}{G-0.02} & \multicolumn{1}{c}{M-20dB} & \multicolumn{1}{c}{M-10dB} & Reverb \\ \midrule
Baseline & \multicolumn{1}{c}{63.2} & \multicolumn{1}{c}{-3.9} & \multicolumn{1}{c}{-7.7} & \multicolumn{1}{c}{-1.7} & \multicolumn{1}{c}{-5.4} & \multicolumn{1}{c|}{-3.9} & \multicolumn{1}{c}{78.7} & \multicolumn{1}{c}{-4.8} & \multicolumn{1}{c}{-9.8} & \multicolumn{1}{c}{-2.2} & \multicolumn{1}{c}{-6.4} & -4.2 \\
Unit & \multicolumn{1}{c}{\textbf{67.9}} & \multicolumn{1}{c}{\textbf{-3.2}} & \multicolumn{1}{c}{\textbf{-6.8}} & \multicolumn{1}{c}{\textbf{-1.4}} & \multicolumn{1}{c}{\textbf{-4.8}} & \multicolumn{1}{c|}{\textbf{-2.6}} & \multicolumn{1}{c}{\textbf{80.9}} & \multicolumn{1}{c}{\textbf{-3.7}} & \multicolumn{1}{c}{\textbf{-5.3}} & \multicolumn{1}{c}{\textbf{-1.5}} & \multicolumn{1}{c}{\textbf{-5.2}} & \textbf{-2.7} \\
\bottomrule
\end{tabular}}
\end{table*}

\subsection{Framework}

The proposed framework is constructed based on the S3PRL toolkit\footnote{https://github.com/s3prl/s3prl} to connect the pre-trained self-supervised model to the downstream model. For the upstream model, we selected HuBERT-Base~\cite{hubert}, and its last layer outputs are used as input for the downstream model. The downstream model is a transformer-based encoder-decoder network designed for the main SLU task. In addition to predicting SLU tasks, we incorporate unit predictions as auxiliary guidance to direct the intermediate representations toward content-related information. The auxiliary unit prediction task shares the same upstream model and encoder as the main task but has a separate transformer decoder for the unit prediction. The overview of the proposed framework is illustrated in Figure \ref{fig:framework}.



During the training phase, the overall loss function is constructed as a weighted sum of the main and auxiliary tasks, and the upstream model is fine-tuned accordingly. The loss function is expressed by the following equation:

\begin{equation}
\begin{aligned}
\mathcal{L} = (1-\lambda) \times \mathcal{L}_{slu} + \lambda \times \mathcal{L}_{aux}
\end{aligned}
\end{equation}
where $\mathcal{L}_{slu}$ is a cross-entropy loss for sequence generation, and $\mathcal{L}_{aux}$ is also a cross-entropy loss but for unit sequence prediction.

\wei{on or for}


\section{Experimental Setup}

\subsection{Dataset}
We evaluate the proposed method on five well-known English SLU datasets: ATIS~\cite{atis},
SLUE-SNER~\cite{slue}, SLURP~\cite{slurp}, SNIPS~\cite{audio_snips}, and STOP~\cite{stop}.
ATIS is a classical corpus that measures the progress of SLU systems with the speech and transcriptions in the aviation domain. SLUE-SNER is a benchmark task created for spoken name entity recognition. We follow the official training set and evaluate using the development set, as the test set is not publicly accessible. SLURP is another widely used SLU package that covers 18 domains and contains a large amount of data. For SNIPS, we use the Slot Filling (SF) task from the SUPERB~\cite{superb} benchmark, which adopts Audio SNIPS~\cite{audio_snips} and follows the standard split in SNIPS~\cite{snips_platform}. Lastly, STOP is a recent SLU semantic parsing dataset that includes a huge amount of audio files and speakers. Overall, we evaluate the SNER task using SLUE-SNER; the ICSF task using ATIS, SLURP, and SNIPS, and the SSP task using STOP.



We measure the performance of the predicted outputs in the main task with several evaluation metrics. \textbf{F1} requires both the slot label and value to be correctly predicted, \textbf{Slot-Type-F1} (ST-F1) measures the F1-score for slot-types, and \textbf{Slot-Value-CER} (SV-CER) calculates the Character Error Rate (CER) for slot-values. \textbf{SLU-F1} is specific to the SLURP dataset and combines F1 evaluation with text-based distance measurements, including CER on the character level and the Word Error Rate (WER). \textbf{INT-Acc} measures the accuracy of predicting intent classes in the ICSF task, and \textbf{EM-Tree} represents the accuracy of producing the semantic parse labels in the SSP task. The corresponding metrics for each dataset are listed in Table \ref{tab:dataset_metric}.


\subsection{Implementation Details}
The experiments are implemented under a 3-layer encoder followed by a 6-layer SLU decoder with a size of 512 for ATIS and 1024 for the rest four datasets. The attention dimension is 512 and there are 4 heads. The unit decoder contains 2 layers and the unit size is 500. For each dataset, there are three ways to train the model. \textbf{Baseline}: only do the main task (i.e., $\lambda=0$). \textbf{Unit}: use discrete units as the target of the auxiliary task with $\lambda=0.5$. \textbf{Text}: use transcripts as the target instead of units. For every experiment, we search the learning rate from the range [$2e-5$, $6e-5$, $2e-4$, $6e-4$, $2e-3$], batch size from [96, 192, 384], and report the best results. The total training steps are 10k. Furthermore, we augment the waveforms of every data with a speed perturbation by a factor of 0.9 and 1.1. 




\begin{table*}[t]
\caption{Results of the performance drop on the test set of SNIPS dataset under different noisy configurations. The notations for noises (G, M, Reverb) are the same as Table \ref{tab:noise_result_slurp}.
}
\vspace{-0.2cm}
\label{tab:noise_result_snips}
\centering
\resizebox{0.98\textwidth}{!}{\begin{tabular}{l|cccccc|cccccc}
\toprule
Metric & \multicolumn{6}{c|}{ST-F1$\uparrow$}                                                                                                                         & \multicolumn{6}{c}{SV-CER$\downarrow$}                                                                                                    \\ \midrule
Noise & \multicolumn{1}{c}{w/o} & \multicolumn{1}{c}{G-0.005} & \multicolumn{1}{c}{G-0.01} & \multicolumn{1}{c}{M-20dB} & \multicolumn{1}{c}{M-10dB} & \multicolumn{1}{c|}{Reverb} & \multicolumn{1}{c}{w/o} & \multicolumn{1}{c}{G-0.005} & \multicolumn{1}{c}{G-0.01} & \multicolumn{1}{c}{M-20dB} & \multicolumn{1}{c}{M-10dB} & Reverb \\ \midrule
Baseline & \multicolumn{1}{c}{77.6} & \multicolumn{1}{c}{-6.9} & \multicolumn{1}{c}{-15.4} & \multicolumn{1}{c}{-3.1} & \multicolumn{1}{c}{-13.4} & \multicolumn{1}{c|}{-2.3} & \multicolumn{1}{c}{42.9} & \multicolumn{1}{c}{+9.1} & \multicolumn{1}{c}{+19.0} & \multicolumn{1}{c}{+4.1} & \multicolumn{1}{c}{+16.2} & -3.3 \\
Unit & \multicolumn{1}{c}{\textbf{82.7}} & \multicolumn{1}{c}{\textbf{-2.9}} & \multicolumn{1}{c}{\textbf{- 9.4}} & \multicolumn{1}{c}{\textbf{-1.9}} & \multicolumn{1}{c}{\textbf{-11.2}} & \multicolumn{1}{c|}{\textbf{-2.1}} & \multicolumn{1}{c}{\textbf{31.9}} & \multicolumn{1}{c}{\textbf{+4.9}} & \multicolumn{1}{c}{\textbf{+14.4}} & \multicolumn{1}{c}{\textbf{+2.8}} & \multicolumn{1}{c}{\textbf{+15.5}} & \textbf{-3.1} \\
\bottomrule
\end{tabular}}
\end{table*}

\section{Results}

Table \ref{tab:main_result} shows the results for ATIS, SLUE-SNER, SLURP, SNIPS, and STOP datasets to compare the baseline approach with our proposed method ``Unit". In Table\ref{tab:main_result}, the row labeled ``Previous" is prior end-to-end SLU works that do not utilize ASR transcripts\footnote{Not every dataset has the results of end-to-end SLU without ASR transcripts, so if some prior works leverage transcripts, we use the notation ``$^*$" to mark the difference.}.

Regarding the \textbf{SNER} problem, the results on the SLUE-SNER dataset demonstrate a 3.8\% improvement in F1-score, 4.0\% in ST-F1, and 5.8\% in SV-CER when unit guidance is utilized. Additionally, the proposed method achieves a comparable F1-score to \cite{slue}, which is a topline approach leveraging paired transcripts and an additional language model for this task. 

For the \textbf{ICSF} tasks, we observed superior performance on the ATIS and SLURP datasets when comparing the "Baseline" with the "Unit" method. For instance, in the case of ATIS, the F1-score improved by 3.3\% and the ST-F1 increased by 1.7\% when unit guidance was added. Furthermore, our method outperforms previous work by 5.8\% and 3.6\% in F1-score and intent accuracy, respectively. Similar results were observed in the SLURP dataset. On the other hand, in the SNIPS dataset, although the previous work has improved the ST-F1 by 7.1\% and the SV-CER by 10.1\% compared to our ``Unit" results, it used additional ASR transcriptions. More importantly, the improvement from ``Baseline" to ``Unit" is about 5.1\% in ST-F1 and 11.0\% in SV-CER, which is a significant increase compared to other datasets.

In the case of the STOP dataset used in the \textbf{SSP} task, we discovered a 4.4\% improvement in the EM-Tree metric when using discrete units, and this improvement is 1.5\% higher than the previous work. It should be noted that while the performance of the previous work is higher than the baseline in our method, we do not use any transcription in the STOP dataset to generate semantic labels. This is a premise that makes the training stage of the SSP task harder than the previous work, and therefore, a direct comparison is not feasible.

To verify the effectiveness of units compared to text, we included the results trained with text guidance (i.e., utilizing the transcripts as targets) in the fourth row labeled "Text". It is worth noting that most of the gaps between "Text" and "Unit" are within the range of 0.1\% to 1.7\%, which is a relatively small performance loss compared to the baseline. These results indicate that our method yields competitive performance to models that leverage paired transcripts.




\section{Discussion}
In the following section, we further explore the advantages of our method, including the few-shot capability, noise robustness, and applicability to other SSL models.
\subsection{Few-Shot Capability}
Table \ref{tab:few_shot_result} displays the training performances of SLURP and SNIPS under few-shot scenarios. Concretely, we randomly sample 10\% of the data from the training and validation sets, while the original test set is utilized for evaluation. All experimental settings are the same as the training process with the full data, except for the number of training epochs, which is reduced to 5k steps during few-shot training. Results demonstrate that utilizing discrete units as guidance leads to a smaller performance drop when the size of the training dataset decreases. Moreover, the performance achieved with units on 10\% SNIPS data is even better than that without units on 100\% SNIPS data, highlighting the few-shot ability of our approach.
\daniel{this paragraph only describes the results. you should give some insights, discussion, etc. about the results.}

We hypothesize that the use of unit guidance for the intermediate representation enables the encoder to capture mostly content information. The SLU decoder benefits from the representative input feature, allowing it to achieve better performance with a limited amount of data.


\subsection{Noise Robustness}
We discover that unit prediction not only guides the encoder to extract content-related information from speech, but also serves as the regularization to help the model focus on useful content instead of irrelevant information (such as additive noises and environmental conditions). In order to discover how unit guidance can help when the target domains contain irrelevant information, we artificially create mismatched test sets with various types of noises. Specifically, noises include additive Gaussian noise (G), background noise MUSAN~\cite{musan} (M), and reverberation effect (Reverb).

In Table \ref{tab:noise_result_slurp} and \ref{tab:noise_result_snips}, we utilize models trained with the setting of ``Baseline" and ``Unit" for the SNIPS and SLURP datasets mentioned in Table \ref{tab:main_result}.  For Gaussian noise, we set the amplitude to 0.005, 0.01, and 0.02. For MUSAN noise, we set the signal-to-noise ratio (SNR) to 20 dB and 10 dB.

As shown in Table \ref{tab:noise_result_slurp} and  \ref{tab:noise_result_snips}, under the same level of domain mismatch, models trained with discrete units on the SLURP and SNIPS have better capacities of noisy robustness and smaller performance degradation in all three mismatched conditions. For instance, when the amplitude of Gaussian noise is set to 0.01, the SNIPS dataset shows a 15.4\% decrease in ``Baseline" ST-F1 and only a 9.4\% decrease in ``Unit" ST-F1. This reveals a 6.0\% difference in performance drop between the two cases, showing better robustness during inference time.

\wei{help converge for SLURP?}



\daniel{similar to section 5.1, you only described results here. some interpretation and discussion about how does your finding inspire future research would be desirable.}

\subsection{Different SSL Models}

In order to explore the applicability of our method to other pre-trained SSL models, we conducted experiments using the wav2vec 2.0~\cite{wav2vec} model while keeping other components consistent. Table \ref{tab:wav_result} presents the results obtained using the wav2vec 2.0 Base model. Our findings from these models, trained on SLURP and SNIPS datasets, show the same trend as observed in Table \ref{tab:main_result}. They suggest that discrete units can serve as effective guidance for different self-supervised models, leading to performance improvements and better training convergence.
\begin{table}[t]
\caption{The performance of unit-guiding and text-guiding approaches using wav2vec 2.0 Base as the upstream SSL model. The dashed line means that the training fails to converge. }
\vspace{-0.2cm}
\label{tab:wav_result}
\centering
\resizebox{0.45\textwidth}{!}{
\begin{tabular}{l|cc|cc}
\toprule
Dataset & \multicolumn{2}{c|}{\textbf{SLURP}}    & \multicolumn{2}{c}{\textbf{SNIPS}}    \\ \midrule
Metric & \multicolumn{1}{c}{SLU-F1$\uparrow$} & INT-Acc$\uparrow$ & \multicolumn{1}{c}{ST-F1$\uparrow$} & SV-CER$\downarrow$ \\ \midrule

Baseline & \multicolumn{1}{c}{--} & -- & \multicolumn{1}{c}{72.2} & 53.0 \\ 
Unit & \multicolumn{1}{c}{\textbf{65.4}} & \textbf{79.5} & \multicolumn{1}{c}{\textbf{78.6}} & \textbf{38.1} \\ \midrule
Text & \multicolumn{1}{c}{66.7} & 80.9 & \multicolumn{1}{c}{80.7} & 35.0 \\

\bottomrule
\end{tabular}}
\end{table}

\section{Conclusion}
In this work, we propose a textless SLU framework that can achieve comparable performance without paired transcripts. We utilize self-supervised discrete units as the intermediate target to guide the model in learning content-related representations. The results show significant improvements accross five SLU benchmarks when compared to the baseline method. Additionally, we demonstrate that our approach provides the few-shot capability (10\% of data) and alleviates the performance degradation in mismatched and noisy environments. Future work includes input augmentation with clean unit guidance to enhance noise robustness and exploring the intermediate target using unsupervised ASR~\cite{baevskiunsupervised, wav2vecu_lin}.
\\
\textbf{Acknowledgement}: We thank the National Center for High-performance Computing (NCHC) of 
National Applied Research Laboratories (NARLabs) in Taiwan for providing computational and storage resources.



\bibliographystyle{IEEEtran}
\bibliography{mybib}

\end{document}